\documentclass[12pt]{article}
\usepackage{amssymb}
\usepackage[margin=1.25 in]{geometry}
\usepackage{amsthm,amsmath,epsfig}
\usepackage{times}
\usepackage[numbers]{natbib}
\usepackage{caption}
\usepackage{color}
\usepackage{bm}
\usepackage{array}
\usepackage{setspace}
\usepackage[ruled,linesnumbered]{algorithm2e}

\begin{document}

{\centering{\Large\bf{Conditional Uncorrelation and Efficient Non-approximate\\ Subset Selection in Sparse Regression}}

\vspace{0.3 cm} {Jianji Wang$^1$, Qi Liu$^{1,2}$, Shupei Zhang$^{1,2}$, Nanning Zheng$^1$, Fei-Yue Wang$^{3}$}

{\small{$^1$Institute of Artificial Intelligence and Robotics, Xi'an
Jiaotong University, Xi'an, China, 710049 \\
$^2$School of Software Engineering, Xi'an
Jiaotong University, Xi'an, China, 710049 \\
$^3$State Key Laboratory of Management and Control
for Complex Systems, Beijing, China, 100190\\
wangjianji@mail.xjtu.edu.cn}}

}

{\center{\bf{Abstract}}

}

Given $m$ $d$-dimensional responsors and $n$ $d$-dimensional predictors, sparse regression finds at most $k$ predictors for each responsor for linear approximation, $1\leq k \leq d-1$. The key problem in sparse regression is subset selection, which usually suffers from high computational cost. Recent years, many improved approximate methods of subset selection have been published. However, less attention has been paid on the non-approximate method of subset selection, which is very necessary for many questions in data analysis. Here we consider sparse regression from the view of correlation, and propose the formula of conditional uncorrelation. Then an efficient non-approximate method of subset selection is proposed in which we do not need to calculate any coefficients in regression equation for candidate predictors. By the proposed method, the computational complexity is reduced from $O(\frac{1}{6}{k^3}+mk^2+mkd)$ to $O(\frac{1}{6}{k^3}+\frac{1}{2}mk^2)$ for each candidate subset in sparse regression. Because the dimension $d$ is generally the number of observations or experiments and large enough, the proposed method can greatly improve the efficiency of non-approximate subset selection.

\vspace{0.2 cm} {\bf{Key words}}: conditional uncorrelation, non-approximate method, sparse regression, subset selection, multivariate correlation.

\vspace{0.2 cm}

{\center\bf{1. \quad INTRODUCTION}

}

The concept of regression, along with the concept of correlation, was firstly discussed by Galton in 1885 \cite{Rodgers1988Thirteen}.
Ten years later, the inner-product correlation coefficient was developed by Pearson. Then the simple regression question can be well solved with the help of Pearson's correlation coefficient \cite{StiglerFrancis}. However, because there was no compact formulation to define the correlation among multiple variables for a long time, people had to analyze the multivariate regression question by the hat-matrix method.

Consider the sparse regression problem with $m$ $d$-dimensional responsors ${\bf{y}}_1, {\bf{y}}_2$, $\cdots$, ${\bf{y}}_m$ and $n$ $d$-dimensional predictors ${\bf{x}}_1, {\bf{x}}_2, \cdots, {\bf{x}}_n$. For each responsor ${\bf{y}}\in\{{\bf{y}}_1, {\bf{y}}_2$, $\cdots, {\bf{y}}_m\}$, at most $k$ predictors are selected to linearly predict ${\bf{y}}$, $1\leq k\leq d-1$. The target of sparse regression is to minimize the $l_2$ distance between ${\bf{y}}$ and the optimum linear combination of the selected predictors for ${\bf{y}}$. Sparse regression problem is usually discussed with a fixed sparsity parameter $k$ \cite{CardinalSparse,Har-Peled2016Approximate}. Without loss of generality, suppose that ${\bf{x}}_1, {\bf{x}}_2, \cdots$, and ${\bf{x}}_k$ are the predictors considered for the responsor ${\bf{y}}$, then
\begin{equation}
	{\bf{y}}=\beta_0{\bf{1}}+\beta_1{\bf{x}}_1+\beta_2{\bf{x}}_2+\cdots+\beta_k{\bf{x}}_k+{\emph{\textbf{e}}},
	\label{eq1}
\end{equation}
where ${\bf{1}}$ is the vector with all ones, ${\emph{\textbf{e}}}$ is the residual, $\beta_i$ is the scalar coefficients of ${\bf{x}}_i$, $i=1,2,\cdots,k$, and $\beta_0$ is the offset. 

In the traditional multivariate regression, we let 
${\bf{X}}$ $=$ $[{\bf{1}},{{\bf{x}}_1},{{\bf{x}}_2}, \cdots ,{{\bf{x}}_k}]$ and ${\bm{\beta}}=[\beta_0,\beta_1,\beta_2,\cdots,\beta_k]^T$, then 
${\bf{y}} = {\bf{X}}{\bm{\beta}} + {\emph{\textbf{e}}}$ and we have
\begin{equation}
	\begin{array}{l}
		\hat{\bm{\beta}} = {({{\bf{X}}^{{T}}}{\bf{X}})^{ - 1}}{{\bf{X}}^{{T}}}{{\bf{y}}} \\
		{\bf{\hat y}} = {\bf{X}}{({{\bf{X}}^{{T}}}{\bf{X}})^{ - 1}}{{\bf{X}}^{{T}}}{{\bf{y}}} \\
		{\emph{\textbf{e}}} = {\bf{y}} - {\bf{X}}({({{\bf{X}}^{{T}}}{\bf{X}})^{ - 1}}({{\bf{X}}^{{T}}}{{\bf{y}}})) \\
	\end{array},
	\label{traditionalequation}
\end{equation}
where $\hat{\bm{\beta}}$ and ${\bf{\hat y}}$ are the estimated vectors of ${\bm{\beta}}$ and ${\bf{y}}$, respectively, and ${\bf{X}}{({{\bf{X}}^{{T}}}{\bf{X}})^{ - 1}}{{\bf{X}}^{{T}}}$ is the hat matrix. The hat-matrix method appeared no later than the 1960s \cite{hoaglin1978hat,tukey1972some}, and has always been an important tool in data analysis.

To select $k$ predictors from the predictor set for a responsor ${\bf{y}}$, we need to consider all the possible subsets with $k$ predictors, which is called subset selection. According to Eq.~(\ref{traditionalequation}), before calculating ${\emph{\textbf{e}}}$ and the $l_2$ norm of ${\emph{\textbf{e}}}$ for subset selection, we have to compute the corresponding linear coefficients. Moreover, the dimension $d$ of the responsors and predictors is generally the number of observations or experiments in statistics and experimental analysis, which is usually large enough. As the size of hat matrix is $d\times d$, the hat-matrix method has a high computational cost.

At present, the applications of sparse regression can be classified into two categories. In the problems belonging to the first category, such as approximate sparse representation, the approximate methods of sparse regression play important roles. For the problems in the second category, like association detection in data, the non-approximate sparse regression is widely used. Many works have been conducted to improve the efficiency of approximate sparse regression. However, the non-approximate subset selection receives little attention in these years. In the era of big data, we urgently need an efficient non-approximate sparse regression method to detect associations among data in large data sets. However, hat-matrix method, which has been used over fifty years, does not meet the needs.

According to Eq.~(\ref{eq1}), sparse regression problem is essentially a correlation problem among the predictors ${\bf{x}}_1, {\bf{x}}_2, \cdots$, ${\bf{x}}_k$ and the responsor ${\bf{y}}$. Here we discuss it from the view of correlation. We have proposed a pair of measures for multivariate correlation, namely, the unsigned correlation coefficient (UCC) $r$ and the unsigned uncorrelation coefficient (UUC) $\omega$ \cite{Wang2014Measures, Wang2018Multivariate}. 
If ${\bf{R}}$ is the correlation matrix of the variables ${\bf{x}}_1, {\bf{x}}_2, \cdots, {\bf{x}}_k$, then $r$ and $\omega$ are defined, respectively, as following:
\begin{equation}
	\begin{array}{l}
		{r^2} = 1 - \det ({\bf{R}}) \\
		{\omega ^2} = \det ({\bf{R}}) \\
	\end{array}.\label{UCCUUC}
\end{equation}
Many important properties and visual figures show that UCC and UUC are the general measures of correlation for multiple variables \cite{Wang2018Multivariate}.

In this paper, the formula of conditional uncorrelation is derived by use of multivariate correlation firstly, then the formula system of the correlation-based multivariate regression is proposed, which can not only speed up the process of regression, but also provide a new way to rethink regression and sparse regression. Based on conditional uncorrelation, we further optimize the calculation of ratio between two determinants of correlation matrices, and propose an efficient method of non-approximate subset selection for sparse regression. Compared with the traditional hat-matrix method, the correlation-based method does not need to compute for linear coefficients of regression equation in subset selection, thereby considerably improving the efficiency of sparse regression.

\vspace{0.2 cm}

{\center\bf{2. \quad CONDITIONAL UNCORRELATION}

}

In sparse regression, it is the target to minimize the mean square error (MSE) between the target vector $\bf{y}$ and its estimated vector $\bf{\hat y}$:
\begin{equation}
	\begin{array}{l}
		{\rm{MSE}}({\bf{y}},{\bf{\hat y}})\! 
		= \!\frac{1}{d}\left\| {{\bf{y}}\! - \!({\beta_0}{\bf{1}}\! + \!{\beta _1}{{\bf{x}}_1}\! + \!{\beta _2}{{\bf{x}}_2}{\rm{ + }} \cdots {\rm{ + }}{\beta _k}{{\bf{x}}_k})} \right\|_2^2
	\end{array}.\label{eq5}
\end{equation}
where $d$ is the dimension of the involved vectors and ${\bf{\hat y}}={\beta_0}{\bf{1}}\! + \!{\beta _1}{{\bf{x}}_1}\! + \!{\beta _2}{{\bf{x}}_2}{\rm{ + }} \cdots {\rm{ + }}{\beta _k}{{\bf{x}}_k}$ is the best linear approximation of $\bf{y}$ by ${\bf{x}}_1, {\bf{x}}_2, \cdots, {\bf{x}}_k$.

Let the standard deviations of the elements in ${\bf{y}}$ and ${\bf{\hat y}}$ are
$\sigma _{\bf{y}}$ and $\sigma _{\bf{\hat y}}$, respectively,
the means of the elements in ${\bf{x}}_i, {\bf{y}}$, and ${\bf{\hat y}}$ are $\mu _i, \mu
_{\bf{y}}$, and $\mu _{\bf{\hat y}}$, respectively, and the covariances
between ${\bf{x}}_i$ and ${\bf{y}}$, between ${\bf{x}}_i$ and
${\bf{x}}_j$, and between ${\bf{y}}$ and ${\bf{\hat y}}$ are $\sigma
_{i{\bf{y}}}$, $\sigma _{ij}$, and $\sigma _{\bf{y{\hat y}}}$,
respectively, $i,j=1,2,\cdots,k$. If ${\bf{x}}=(x_1,x_2,\cdots,x_d)^{{T}}$ and ${\bf{y}}=(y_1,y_2,\cdots,y_d)^{{T}}$, here the variance of the elements in ${\bf{x}}$ is defined as $\sigma^2_{\bf{x}}=\frac{1}{d}{\sum\limits_{i=1}^n}(x_i-\mu_{\bf{x}})^2$, the standard deviation of the elements in ${\bf{x}}$ is defined as $\sigma_{\bf{x}}=\sqrt{\sigma^2_{\bf{x}}}$, 
and the covariance between ${\bf{x}}$ and ${\bf{y}}$ is defined as $\sigma _{\bf{xy}}=\frac{1}{d}{\sum\limits_{i=1}^n}(x_i-\mu_{\bf{x}})(y_i-\mu_{\bf{y}})$. 

Then we have the first lemma as follows:

{\noindent\bf{Lemma 1.}} For a responsor ${\bf{y}}$ and $k$ linearly independent predictors ${{\bf{x}}_1},{{\bf{x}}_2}{\rm{,}} \cdots{\rm{,}}{{\bf{x}}_k}$, suppose the best linear approximation of ${\bf{y}}$ by the $k$ predictors is ${\hat{\bf{y}}}$, ${\hat{\bf{y}}}={\hat{\beta}_0}{\bf{1}}\! + \!{\hat{\beta}_1}{{\bf{x}}_1}\! + \!{\hat{\beta}_2}{{\bf{x}}_2}{\rm{ + }} \cdots {\rm{ + }}{\hat{\beta}_k}{{\bf{x}}_k}$. We denote by ${\omega({{\bf{x}}_1},{{\bf{x}}_2}{\rm{,}} \cdots{\rm{,}}{{\bf{x}}_k})}$ the UUC among the $k$ predictors, and denote by ${\omega({{\bf{x}}_1},{{\bf{x}}_2}{\rm{,}} \cdots{\rm{,}}{{\bf{x}}_k},{\bf{y}})}$ the UUC among the $k+1$ variables ${{\bf{x}}_1},{{\bf{x}}_2}{\rm{,}} \cdots{\rm{,}}{{\bf{x}}_k},{\bf{y}}$. Let the variance of the elements in ${\bf{y}}$ is $\sigma _{\bf{y}}^2$, then we have

\begin{equation}
	{\rm{MSE}} ({\bf{y},\bf{\hat y}}) = \sigma _{\bf{y}}^2\cdot \displaystyle\frac{{{\omega ^2}({{\bf{x}}_1},{{\bf{x}}_2}{\rm{,}} \cdots{\rm{,}}{{\bf{x}}_k},{\bf{y}})}}{{{\omega ^2}({{\bf{x}}_1},{{\bf{x}}_2}{\rm{,}} \cdots {\rm{,}}{{\bf{x}}_k})}}.
	\label{eq6}
\end{equation}

\begin{proof}
	To minimize the value of MSE between ${\bf{y}}$ and ${\bf{\hat y}}$ by
	the least square method, we have
	\begin{equation}
		\begin{array}{*{20}{l}}
			{{\mu _{\bf{y}}} = {\mu _{\bf{\hat y}}} = {{\hat \beta }_0}+{{\hat \beta }_1}{\mu _1} + {{\hat \beta }_2}{\mu _2} +  \cdots  + {{\hat \beta }_k}{\mu _k}}  \\
			{\sum\limits_j {{{\hat \beta }_j}{\sigma _{ij}}}  = {\sigma _{i{\bf{y}}}}}  \\
			{\rm{MSE}} ({\bf{y}},{\bf{\hat y}}) = \sigma_{\bf{y}}^2  - 2\sigma _{{\bf{y\hat y}}}  +\sigma _{\bf{\hat y}}^2 \\
		\end{array},\label{eq7}
	\end{equation}
	$i=1,2,\cdots,k$. According to the definitions of variance and covariance, we have
	\begin{equation}
		\begin{array}{l}
			\sigma _{\bf{\hat y}}^2  = \sum\limits_i {\hat{\beta}_i \sum\limits_j {\hat{\beta}_j \sigma _{ij} } }  \\
			\sigma _{{\bf{y\hat y}}}  = \sum\limits_i {\hat{\beta}_i \sigma _{i{\bf{y}}} }  \\
		\end{array}.\label{eq4}
	\end{equation}
	
	Then we have the relation between $\sigma _{\bf{{\hat y}}}^2$ and $\sigma_{{\bf{y\hat y}}}$:
	\begin{equation}
		\sigma _{\bf{{\hat y}}}^2 = \sum\limits_i {\hat\beta _i \sum\limits_j {\hat\beta _j \sigma _{ij} }} = \sum\limits_i {\hat\beta _i {\sigma _{i{\bf{y}}}} } = \sigma_{{\bf{y\hat y}}}.\label{eq9}
	\end{equation}
	
	Let ${\bm\sigma}\!\!=\!\!\left[\!\!{\begin{array}{*{20}c}
			{\sigma _{11} } & {\sigma _{12} } &  \cdots  & {\sigma _{1k} }  \\
			{\sigma _{21} } & {\sigma _{22} } &  \cdots  & {\sigma _{2k} }  \\
			\vdots  &  \vdots  &  \ddots  &  \vdots   \\
			{\sigma _{k1} } & {\sigma _{k2} } &  \cdots  & {\sigma _{kk} }  \\
	\end{array}} \!\!\right]\!$,
	${\hat{\bm{\beta}}}\!=\!\!\left[\! {\begin{array}{*{20}c}
			{\hat \beta_1 }  \\
			{\hat \beta_2 }  \\
			\vdots   \\
			{\hat \beta_k }  \\
	\end{array}}\! \right]\!$, 
	$ {\bm\sigma}_{ \bullet {\bf{y}}}\!\!=\!\!\left[\!\! {\begin{array}{*{20}c}
			{\sigma _{1{\bf{y}}} }  \\
			{\sigma _{2{\bf{y}}} }  \\
			\vdots   \\
			{\sigma _{k{\bf{y}}} }  \\
	\end{array}}\! \right]\!$. 
	According to the second equation in Eq.~(\ref{eq7}), we have
	\begin{equation}
		{\hat{\bm{\beta}}}={\bm\sigma }^{ - 1}{\bm\sigma}_{\bullet {\bf{y}}}. \label{eq10}
	\end{equation}
	
	Combining Eqs.~(\ref{eq9}) and (\ref{eq10}) we can obtain \cite{Wang2021}
	\begin{equation}
		\sigma _{{\bf{y\hat y}}}={{\bm\sigma}^{T}_{\bullet {\bf{y}}}}{\bm\sigma}^{ - 1} {\bm\sigma}_{\bullet {\bf{y}}}.
	\end{equation}
	
	Moreover,
	\begin{equation}
		\det \left[ {\begin{array}{*{20}c}
				{\bm\sigma} & {{\bm\sigma}_{\bullet {\bf{y}}}}  \\
				{{\bm\sigma}^T_{\bullet {\bf{y}}} } & {\sigma _{\bf{y}}^2 }  \\
		\end{array}} \right] = \det ({\bf{\bm\sigma}})(\sigma _{\bf{y}}^2  - {{\bm\sigma}^T_{\bullet {\bf{y}}}} {\bf{\bm\sigma }}^{ - 1} {\bm\sigma}_{\bullet {\bf{y}}} ).\label{para}
	\end{equation}
	
	Hence,
	\begin{equation}
		\begin{array}{*{20}l}
			{\rm{MSE}} ({\bf{y},\bf{\hat y}}) = \sigma_{\bf{y}}^2  - \sigma_{{\bf{y\hat y}}} 
			= \displaystyle\frac{\det \left[ {\begin{array}{*{20}c}
						{\bm\sigma} & {{\bm\sigma}_{\bullet {\bf{y}}}}  \\
						{{\bm\sigma}^T_{\bullet {\bf{y}}}} & {\sigma _{\bf{y}}^2 }  \\
				\end{array}} \right]}{\det ({\bf{\bm\sigma}})} \\
			{ = \sigma _{\bf{y}}^2\cdot \displaystyle\frac{{{\omega ^2}({{\bf{x}}_1},{{\bf{x}}_2}{\rm{,}} \cdots{\rm{,}}{{\bf{x}}_k},{\bf{y}})}}{{{\omega ^2}({{\bf{x}}_1},{{\bf{x}}_2}{\rm{,}} \cdots {\rm{,}}{{\bf{x}}_k})}}}.
		\end{array}
		\label{eq13}
	\end{equation}
\end{proof}

Let $r({{\bf{y}},{\bf{\hat y}})}$ and $\omega({{\bf{y}},{\bf{\hat y}})}$ are the unsigned bivariate correlation coefficient (absolute value of Pearson's correlation coefficient) and the unsigned bivariate uncorrelation coefficient between ${\bf{y}}$ and ${\bf{\hat y}}$, respectively, then
\[
r^2({{\bf{y,\hat y}}}) \! = \!(\frac{{\sigma
		_{{\bf{y\hat y}}} }}{{\sigma _{\bf{y}} \sigma _{\bf{\hat y}} }})^2 \! =\!
\frac{{\sigma _{{\bf{y\hat y}}}
}}{{\sigma _{\bf{y}}^2 }}\! = \!1 - \!\frac{{{\omega ^2}({{\bf{x}}_1},{{\bf{x}}_2}{\rm{,}} \cdots {\rm{,}}{{\bf{x}}_k},{\bf{y}})}}{{{\omega ^2}({{\bf{x}}_1},{{\bf{x}}_2}{\rm{,}} \cdots {\rm{,}}{{\bf{x}}_k})}}.
\]

Therefore,
\begin{equation}
	\omega ({\bf{y}},{\bf{\hat y}}) = \frac{{\omega ({{\bf{x}}_1},{{\bf{x}}_2}{\rm{,}} \cdots {\rm{,}}{{\bf{x}}_k},{\bf{y}})}}{{\omega ({{\bf{x}}_1},{{\bf{x}}_2}{\rm{,}} \cdots {\rm{,}}{{\bf{x}}_k})}}.\label{eq15}
\end{equation}

Here $\omega ({\bf{y}},{\bf{\hat y}})$ measures the uncorrelation between ${\bf{y}}$ and the optimum linear approximation constructed by ${\bf{x}}_1, {\bf{x}}_2, \cdots$, and ${\bf{x}}_k$ for ${\bf{y}}$. We can thus use $\omega ({\bf{y}} | {{\bf{x}}_1},{{\bf{x}}_2}{\rm{,}} \cdots {\rm{,}}{{\bf{x}}_k})$ to denote $\omega ({\bf{y}},{\bf{\hat y}})$. Then
\begin{equation}
	\omega ({\bf{y}} | {{\bf{x}}_1},{{\bf{x}}_2}{\rm{,}} \cdots {\rm{,}}{{\bf{x}}_k}) = \frac{{\omega ({{\bf{x}}_1},{{\bf{x}}_2}{\rm{,}} \cdots {\rm{,}}{{\bf{x}}_k},{\bf{y}})}}{{\omega ({{\bf{x}}_1},{{\bf{x}}_2}{\rm{,}} \cdots {\rm{,}}{{\bf{x}}_k})}},\label{conditionaluncorrerlation}
\end{equation} 

We call Eq.~(\ref{eq15}) and Eq.~(\ref{conditionaluncorrerlation}) as {\bf{the formula of conditional uncorrelation}}, which offers the relation among the UUC between the target vector ${\bf{y}}$ and its estimated vector ${\bf{\hat y}}$, the UUC among these vector ${\bf{x}}_1 ,{\bf{x}}_2 , \cdots$, ${\bf{x}}_m,
{\bf{y}}$, and the UUC among these predictor vectors ${\bf{x}}_1 ,{\bf{x}}_2 ,
\cdots ,{\bf{x}}_m$.

For any responsor ${\bf{y}}$, $\sigma _{\bf{y}}^2$ is kept invariant. Then it comes to an interesting conclusion:

{\noindent\bf{Corollary 1.}} To select the best subset for any responsor ${\bf{y}}$ in sparse regression, we have
\begin{equation}
	\mathop{\min}\limits_{{{\bf{x}}_{1}},{{\bf{x}}_{2}}, \cdots ,{{\bf{x}}_{k}}} {\rm{MSE}}({\bf{y}},{\bf{\hat y}})
	\Leftrightarrow \mathop{\min}\limits_{{{\bf{x}}_{1}},{{\bf{x}}_{2}}, \cdots ,{{\bf{x}}_{k}}} \frac{{\omega^2 ({{\bf{x}}_1},{{\bf{x}}_2}{\rm{,}} \cdots {\rm{,}}{{\bf{x}}_k},{\bf{y}})}}{{\omega^2 ({{\bf{x}}_1},{{\bf{x}}_2}{\rm{,}} \cdots {\rm{,}}{{\bf{x}}_k})}},
	\label{eq14}
\end{equation}
where ${\bf{\hat y}}$ is the best linear approximation of ${\bf{y}}$ by ${{\bf{x}}_{1}}$, ${{\bf{x}}_{2}}$, $\cdots$,  ${{\bf{x}}_{k}}$.

The above conclusion provides the target function to choose predictors for a responsor ${\bf{y}}$ in the conditional uncorrelation-based sparse regression. It shows from Eqs.~(\ref{eq15}) and (\ref{eq14}) that the minimization of
${\rm{MSE}}({\bf{y}},{\bf{\hat y}})$ is equivalent to minimizing the conditional uncorrelation between ${\bf{y}}$ and ${\bf{\hat y}}$, which is kept the same with the univariate case \cite{Wang2013A}.

Additionally, the coefficient of multiple determination $R$
\cite{Kutner2005Applied} can also be simplified as
\begin{equation}
	R^2  = \frac{{SSR}}{{SSTO}} = \frac{{\sigma _{\bf{\hat y}}^2 }}{{\sigma _{\bf{y}}^2 }} =
	\frac{{{\sigma_{{\bf{y\hat y}}}}
	}}{{\sigma _{\bf{y}}^2 }} = r^2({{\bf{y,\hat y}}}).\label{eq16}
\end{equation}

According to Eqs. (\ref{eq14}) and (\ref{eq16}), we have
\begin{equation}
	\min {\rm{MSE}}({\bf{y}},{\bf{\hat y}})
	\Leftrightarrow \max R^2 \Leftrightarrow \min \omega^2({\bf{y}},{\bf{\hat y}}).
\end{equation}

Therefore, in sparse regression, it is consistent to select the predictors for a responsor ${\bf{y}}$ whether the target is to minimize MSE between $\bf{y}$ and $\bf{\hat y}$, to minimize UUC between $\bf{y}$ and $\bf{\hat y}$, or to maximize the coefficient of multiple determination.

Lastly, by minimizing the conditional uncorrelation between ${\bf{y}}$ and ${\bf{\hat y}}$ we can select the best subset with $k$ predictors for $\bf{y}$ according to Eq. $\!$(\ref{eq14}). The linear coefficients of the selected predictors in the best subset can be computed by the following equation, which can be easily derived from Eq. (\ref{eq10}): 
\begin{equation}
	\left( {\begin{array}{*{20}{c}}
			{{\sigma _1}\hat{\beta} _1}  \\
			{{\sigma _2}\hat{\beta} _2}  \\
			\vdots   \\
			{{\sigma _k}\hat{\beta} _k}  \\
	\end{array}} \right){\rm{ = }}{\sigma _{\bf{y}}}{{\bf{R}}^{ - 1}_{\bf{x}}}\left( {\begin{array}{*{20}{c}}
			{{\rho_{1{\bf{y}}}}}  \\
			{{\rho_{2{\bf{y}}}}}  \\
			\vdots   \\
			{{\rho_{k{\bf{y}}}}}  \\
	\end{array}} \right)
	\label{eq18}
\end{equation}
where ${\bf{R_x}}$ is the correlation matrix of ${{\bf{x}}_1},{{\bf{x}}_2}{\rm{,}} \cdots {\rm{,}}{{\bf{x}}_k}$, and $\sigma_i$ is the standard deviation of the elements in ${\bf{x}}_i$, $i=1,2,\cdots,k$. 
Then the formula system of the correlation-based multivariate regression is composed of Eqs. (\ref{eq6}), (\ref{eq14}), and (\ref{eq18}).

As discussed above, linear regression is essentially a correlation problem among predictors and responsors. Here we study multivariate regression from the view of correlation, and propose a new formula system of regression. Similar to the hat-matrix-based formula system as shown in Eq.~\ref{traditionalequation}, the formula system of the correlation-based multivariate regression is also a non-approximate method of regression, which can provide with a new way to rethink regression and sparse regression. For example, some approximate strategies may be applied to the proposed scheme by taking correlation as a feature to further improve the efficiency of sparse regression. In the following sections, we will try to improve the efficiency of sparse regression based on the formula of conditional uncorrelation.

\vspace{2mm}

{\center\bf{3. \quad EFFICIENT SUBSET SELECTION IN SPARSE REGRESSION}

}

In sparse regression, we select the best subset for ${\bf{y}}$ to minimize $l_2$ norm of the residual vector ${\emph{\textbf{e}}}$. For each subset, the time complexity $T$ to compute $l_2$ norm of ${\emph{\textbf{e}}}$ in the hat-matrix method is related to both $d$ and $k$ so that $T=T(d,k)$ according to Eq. (\ref{traditionalequation}). In the conditional uncorrelation-based sparse regression, we need to compute the ratio of two determinants of correlation matrices for each subset according to Eqs. (\ref{UCCUUC}) and (\ref{eq14}) so that $T=T(k)$. As we use $k$ predictors and the vector $\bf{1}$ to linearly approximate each $d$-dimensional responsor in sparse regression, when $k=d-1$ the responsors can be losslessly represented if these predictors are linearly independent. Generally, we have $k\ll d-1$. Hence, the correlation-based sparse regression may be more efficient than the traditional hat-matrix method.

\vspace{2mm}
{\noindent\bf{3.1 Method}}

In the proposed method, the best subset for a responsor ${\bf{y}}$ can be selected by Eq. (\ref{eq14}). We denote by ${\bf{R_{x}}}$ the correlation matrix of ${\bf{x}}_1, {\bf{x}}_2, \cdots$, ${\bf{x}}_k$, denote by ${\bf{R_{xy}}}$ the correlation matrix of ${{\bf{x}}_1},{{\bf{x}}_2},\cdots,{{\bf{x}}_k},{{\bf{y}}}$, and let the correlation coefficients related to {\bf{y}} lie in the last row and last column of ${\bf{R_{xy}}}$. According to the definition of UUC in Eq. (\ref{UCCUUC}), we need to minimize the ratio between the determinants of ${\bf{R_{xy}}}$ and ${\bf{R_{x}}}$. 

In practice, we first calculate all the correlation coefficients between two predictors and all the correlation coefficients between the responsor and each predictor. Then in the process of subset selection, ${\bf{R_{xy}}}$ and ${\bf{R_{x}}}$ can be directly constructed by the calculated correlation coefficients .

The main computational cost of sparse regression comes from subset selection. Compared with subset selection, the computational cost of other parts, such as calculation of all correlation coefficients, is negligible.

Here we calculate the ratio between the determinants of ${\bf{R_{xy}}}$ and ${\bf{R_{x}}}$ by upper triangulation of matrix. After upper triangulation of a matrix by adding a scalar multiple of front row to rear row, the determinant of the original matrix is equal to the product of all the diagonal elements in the triangularized matrix of the original matrix. Because the product of the diagonal elements of triangularized ${\bf{R_x}}$ is kept the same with the product of the first $k$ diagonal elements of triangularized ${\bf{R_{xy}}}$ by adding a scalar multiple of front row to rear row, the last diagonal element of triangularized ${\bf{R_{xy}}}$ is just $\omega^2({\bf{y}},{\bf{\hat y}})$. Hence, we can compute $\omega^2 ({\bf{y}},{\bf{\hat y}})$ according to Algorithm~\ref{alg:a1}.

\begin{algorithm}[t]
	\DontPrintSemicolon 
	\KwIn{Correlation matrix ${\bf{R_{xy}}}$ of ${{\bf{x}}_1},{{\bf{x}}_2}{\rm{,}} \cdots {\rm{,}}{{\bf{x}}_k},{\bf{y}}$}
	\KwOut{$\omega^2 ({\bf{y}},{\bf{\hat y}})$}
	\For{$i \gets 1$ \textbf{to} $k$} {
		recipdiag = $1/{\bf{R_{xy}}}[i][i]$ \\
		\For{$j \gets i\!+\!1$ \textbf{to} $k\!+\!1$}{
			temp $= {\bf{R_{xy}}}[i][j]*$recipdiag;\\
			\For{$p \gets j$ \textbf{to} $k\!+\!1$}{
				${\bf{R_{xy}}}[j][p] = {\bf{R_{xy}}}[j][p]-{\bf{R_{xy}}}[i][p]*$temp;
			}
		}
	}
	\Return{${\bf{R_{xy}}}[k+1][k+1]$}
	\caption{Calculation of $\omega^2 ({\bf{y}},{\bf{\hat y}})$}
	\label{alg:a1}
\end{algorithm}

Two strategies are used in Algorithm~\ref{alg:a1} to improve the efficiency of algorithm. Firstly, we ignore all the numbers which have no effect on the calculation of diagonal elements. By this strategy, $p$ gets the values from $i+1$ to $k+1$ in Line 6 in Algorithm~\ref{alg:a1}, and the value $i$ for $p$ can be ignored. Secondly, we optimize the algorithm by the symmetry of correlation matrix. We use ${\bf{R_{xy}}}[i][j]$ instead of ${\bf{R_{xy}}}[j][i]$ in Line 4 of Algorithm~\ref{alg:a1}. Then $p$ can be only discussed from $j$ to $k+1$, and the values from $i+1$ to $j-1$ for $p$ can be ignored by considering both strategies.

\vspace{2mm}
{\noindent\bf{3.2 Algorithm Optimization}}

Here we further optimize the algorithm shown in Algorithm~\ref{alg:a1}. Obviously, for sparse regression with $m$ responsors, Algorithm~\ref{alg:a1} needs to be performed $m$ times for each candidate subset to select the best subsets for all responsors. 

According to the definition of correlation matrix, the first element in correlation matrix is always 1. Then the value of `recipdiag' in Algorithm~\ref{alg:a1} is also 1 for the first row of ${\bf{R_{xy}}}$. Hence, we do not need to perform the multiplications for the codes in Line 4 of Algorithm~\ref{alg:a1} when $i=1$, which can reduce 1 division and $k$ multiplications. This is also why we use correlation matrices in the proposed algorithm although it can be transformed into other forms such as inner product matrices and covariance matrices.

Moreover, let ${{\bf{R}}_{\bullet{\bf{y}}}} = {[{\rho _{1{\bf{y}}}},{\rho _{2{\bf{y}}}}, \cdots ,{\rho _{k{\bf{y}}}}]^T}$ where ${\rho _{i{\bf{y}}}}$ is Pearson's correlation coefficient between ${\bf{x}}_i$ and ${\bf{y}}$, $i=1,2$, $\cdots,k$, then we have 
\[{{\bf{R}}_{{\bf{xy}}}} = \left[ {\begin{array}{*{20}{c}}{{{\bf{R}}_{\bf{x}}}} & {{{\bf{R}}_{ \bullet {\bf{y}}}}}  \\	{{\bf{R}}_{ \bullet {\bf{y}}}^T} & 1  \\ \end{array}} \right].\]
Obviously, when the matrix ${{\bf{R}}_{{\bf{xy}}}}$ is triangularized by adding a scalar multiple of front row to rear row and by multiplying a row with a nonzero constant, we can perform exactly the same operations on ${{\bf{R}}_{{\bf{x}}}}$ for different responsors. Therefore, the upper triangular matrix of ${{\bf{R}}_{{\bf{x}}}}$ can be reused to further improve the efficiency of the proposed algorithm. 

Then we have the following lemma:

{\noindent\bf{Lemma 2.}} Let $\eta_i=$ recipdiag with index $i$ in Algorithm~\ref{alg:a1}. Suppose the upper triangular matrix of ${{\bf{R}}_{{\bf{x}}}}$ by performing the operations of adding a scalar multiple of front row to rear row and the operations of multiplying a row with a nonzero constant is as following:
\[{{\bf{R}}_{\bf{x}}}\xrightarrow{\rm{Upper\ Triangulation}} {\bf{R}}_t=\left[ {\begin{array}{*{20}{c}}
		1 & {{a_{12}}} & {{a_{13}}} &  \cdots  & {{a_{1k}}}  \\
		0 & 1 & {{a_{23}}} &  \cdots  & {{a_{2k}}}  \\
		0 & 0 & 1 &  \cdots  & {{a_{3k}}}  \\
		\vdots  &  \vdots  &  \vdots  &  \ddots  &  \vdots   \\
		0 & 0 & 0 &  \cdots  & 1  \\
\end{array}} \right]\!.\]
Similarly, by the exactly same operations in Rows 1 to $k$ and only by use of the operations of adding a scalar multiple of front row to $k+1$ row in ${{\bf{R}}_{\bf{xy}}}$, we have
\begin{equation}
	{{\bf{R}}_{\bf{xy}}}\xrightarrow{\rm{Upper\ Triangulation}} \left[ {\begin{array}{*{20}{c}}
			{\bf{R}}_t  & {\bf{f}}  \\
			{\bf{0}} &  \omega^2 ({\bf{y}},{\bf{\hat y}}) \\
	\end{array}} \right],
	\label{kTri}
\end{equation}
where ${\bf{f}}=\left[b_1\eta_1,b_2\eta_2,\cdots,b_k\eta_k\right]^T$, then we have
\begin{equation}
	\begin{array}{l}
		{b_i} = {\rho _{i{\bf{y}}}} - \sum\limits_{j = 1}^{i - 1} {{a_{ji}}{b_j}}  \\ 
		{\omega ^2}({\bf{y}},{\bf{\hat y}}) = 1 - \sum\limits_{j = 1}^k {b_j^2{\eta _j}}  \\ 
	\end{array},\label{imp}
\end{equation}
where $b_1={\rho _{1{\bf{y}}}}$ and $i=2,3,\cdots,k$.

\begin{proof}
	We prove Eq.~(\ref{imp}) by mathematical induction. 
	
	When $k=1$, we have $b_1=\rho _{1{\bf{y}}}$ and
	\[
	{\omega ^2}({\bf{y}},{\bf{\hat y}}) = 1 - b_1^2\eta_1 = 1 - \rho _{1{\bf{y}}}^2 = \frac{{\det \left[ {\begin{array}{*{20}{c}}
					1 & {{\rho _{1{\bf{y}}}}}  \\
					{{\rho _{1{\bf{y}}}}} & 1  \\
			\end{array}} \right]}}{{\det \left[ 1 \right]}}.
	\]
	
	Therefore, Eq.~(\ref{imp}) is true when $k=1$.
	
	Suppose Eq.~(\ref{imp}) is true when $k=n$.
	
	When $k=n+1$, the same operations in Lemma 2 are performed on ${{\bf{R}}_{{\bf{xy}}}}$ from Rows 1 to $n$, and only the operations of adding a scalar multiple of front row to the last row are performed on the last row of ${{\bf{R}}_{{\bf{xy}}}}$, then we have
	\begin{equation}
		\begin{array}{l}
			\left[ {\begin{array}{*{20}{c}}
					1 & {{\rho _{12}}} &  \cdots  & {{\rho _{1n}}} & {{\rho _{1,n + 1}}} & {{\rho _{1{\bf{y}}}}}  \\
					{{\rho _{21}}} & 1 &  \cdots  & {{\rho _{2n}}} & {{\rho _{2,n + 1}}} & {{\rho _{2{\bf{y}}}}}  \\
					\vdots  &  \vdots  &  \ddots  &  \vdots  &  \vdots  &  \vdots   \\
					{{\rho _{n1}}} & {{\rho _{n2}}} &  \cdots  & 1 & {{\rho _{n,n + 1}}} & {{\rho _{n{\bf{y}}}}}  \\
					{{\rho _{n + 1,1}}} & {{\rho _{n + 1,2}}} &  \cdots  & {{\rho _{n + 1,n}}} & 1 & {{\rho _{n + 1,{\bf{y}}}}}  \\
					{{\rho _{1{\bf{y}}}}} & {{\rho _{2{\bf{y}}}}} &  \cdots  & {{\rho _{n{\bf{y}}}}} & {{\rho _{n + 1,{\bf{y}}}}} & 1  \\
			\end{array}} \right]\vspace{1ex}  \\
			
			\xrightarrow[\rm{Triangulation}]{\rm{Upper}}\left[ {\begin{array}{*{20}{c}}
					1 & {{a_{12}}} &  \cdots  & {{a_{1n}}} & {{a_{1,n + 1}}} & {{c_1\eta_1}}  \\
					0 & 1 &  \cdots  & {{a_{2n}}} & {{a_{2,n + 1}}} & {{c_2\eta_2}}  \\
					\vdots  &  \vdots  &  \ddots  &  \vdots  &  \vdots  &  \vdots   \\
					0 & 0 &  \cdots  & 1 & {{a_{n,n + 1}}} & {{c_n\eta_n}}  \\
					0 & 0 &  \cdots  & 0 & 1 & {{c_{n + 1}\eta_{n+1}}}  \\
					0 & 0 &  \cdots  & 0 & 0 & \omega^2({\bf{y}},{\bf{\hat y}})  \\
			\end{array}}\! \right]\!\!.\vspace{0.5ex}
		\end{array}\label{k1Tri}
	\end{equation}
	Obviously, $a_{ij}, i\!<\!j\!\le\! n$, is kept the same with $a_{ij}$ in the case of $k=n$, and $\eta_i$ is also kept the same with $\eta_i$ in the case of $k=n$, $i=1,2,\cdots,n$. Moreover, we have ${a_{1,n + 1}}={\rho_{1,n + 1}}$ and $c_i=b_i$ for $i=1,2,\cdots,n$. Here we can take ${\bf{x}}_{k+1}$ as a responsor in Eq.~(\ref{kTri}). Then according to the results of Eq.~(\ref{imp}) in case of $k=n$, we have the following results for Eq.~(\ref{k1Tri}), $i=2,3,\cdots,n$:
	\begin{equation}
		\begin{array}{l}
			\frac{a_{i,n+1}}{\eta_i} = {\rho_{i,n+1}} - \sum\limits_{j = 1}^{i - 1} {a_{ji}}\frac{{a_{j,n+1}}}{\eta_j}  \\ 
			\eta_{n+1} = \frac{1}{1 - \sum\limits_{j = 1}^n  {(\frac{a_{j,n+1}}{{\eta _j}})}^2{{\eta _j}}}  \\ 
		\end{array}.
		\label{eq20}
	\end{equation}
	
	Then only $c_{n+1}$ and $\omega^2({\bf{y}},{\bf{\hat y}})$ are not calculated. At this time, ${{\bf{R}}_{{\bf{xy}}}}$ with $k=n+1$ is transformed to
	\[
	\left[ {\begin{array}{*{20}{c}}
			1 & {{a_{12}}} &  \cdots  & {{a_{1n}}} & {{a_{1,n + 1}}} & {{c_1\eta_1}}  \\
			0 & 1 &  \cdots  & {{a_{2n}}} & {{a_{2,n + 1}}} & {{c_2\eta_2}}  \\
			\vdots  &  \vdots  &  \ddots  &  \vdots  &  \vdots  &  \vdots   \\
			0 & 0 &  \cdots  & 1 & {{a_{n,n + 1}}} & {{c_n\eta_n}}  \\
			{{\rho _{1,n + 1}}} & {{\rho _{2,n + 1}}} &  \cdots  & {{\rho _{n,n + 1}}} & 1 & {{\rho _{n + 1,{\bf{y}}}}}  \\
			{{\rho _{1{\bf{y}}}}} & {{\rho _{2{\bf{y}}}}} &  \cdots  & {{\rho _{n{\bf{y}}}}} & {{\rho _{n + 1,{\bf{y}}}}} & 1  \\
	\end{array}} \right].
	\]
	
	By the operations we used here, we can easily obtain that if 
	\begin{equation}
		\begin{array}{l}
			{d_i} = {\rho _{i,n + 1}} - \sum\limits_{j = 1}^{i - 1} {{a_{ji}}{d_j}}  \\
			{e_i} = {\rho _{i{\bf{y}}}} - \sum\limits_{j = 1}^{i - 1} {{a_{ji}}{e_j}} 
		\end{array}
		\label{eq21}
	\end{equation}
	then 
	\begin{equation}
		\begin{array}{l}
			{c_{n + 1}} = {\rho _{n + 1,{\bf{y}}}} - \sum\limits_{j = 1}^n {{c_j}{\eta _j}{d_j}}  \\
			{\omega ^2} = 1 - \sum\limits_{j = 1}^{n + 1} {{c_j}{\eta _j}{e_j}} 
		\end{array}
		\label{eq22}.
	\end{equation}
	
	Compare the first equation in Eq.~(\ref{eq21}) with the first equation in Eq.~(\ref{eq20}), and compare the second equation in Eq.~(\ref{eq21}) with the first equation in Eq.~(\ref{imp}), then we can obtain $d_j=\frac{a_{j,n+1}}{\eta_j}$ and $e_j=c_j$. Substitute them into Eq.~(\ref{eq22}) we have
	\[
	\begin{array}{l}
		{c_{n + 1}} = {\rho _{n + 1,{\bf{y}}}} - \sum\limits_{j = 1}^n {{a_{j,n+1}}{c_j}}  \\
		{\omega^2({\bf{y}},{\bf{\hat y}})} = 1 - \sum\limits_{j = 1}^{n + 1} {{c_j^2}{\eta _j}} 
	\end{array},
	\]
	which also accord with Eq.~(\ref{imp}). 
	
	Hence, Eq.~(\ref{imp}) is true when $k=n+1$.
	
\end{proof}

According to Lemma 2, we have the optimized method provided in Algorithm~\ref{alg:a2}. If there is only one responsor, Algorithm~\ref{alg:a2} can be directly used. If there are multiple responsors, only the codes in Lines 16 to 25 need to be executed for each responsor, and the codes in Line 1 to 15 need to be executed only once.

In Algorithm~\ref{alg:a2}, we can save eta$[i]$ in ${\bf{R}}_t[i][1]$ to save spaces. In fact, it needs a total of $\frac{(k-1)(k+2)}{2}$ units of space to save eta and ${\bf{R}}_t$.

\begin{algorithm}[!ht]
	\DontPrintSemicolon
	\KwIn{Correlation matrix ${\bf{R_{x}}}$ of ${{\bf{x}}_1},{{\bf{x}}_2}{\rm{,}} \cdots {\rm{,}}{{\bf{x}}_k}$;  \\
		\hspace{9.6mm} ${{\bf{R}}_t} = \left[ {\begin{array}{*{20}{c}}
				1 & {{\rho _{12}}} & {{\rho _{13}}} &  \cdots  & {{\rho _{1k}}}  \\
				0 & 1 & 0 &  \cdots  & 0  \\
				0 & 0 & 1 &  \cdots  & 0  \\
				\vdots  &  \vdots  &  \vdots  &  \ddots  &  \vdots   \\
				0 & 0 & 0 &  \cdots  & 1  \\
		\end{array}} \right]$; \\
		\hspace{9.6mm} Correlation coefficient $\rho_{i{\bf{y}}}$ between ${\bf{x}}_i$ and ${\bf{y}}$; 	}	
	\KwOut{$\omega^2 ({\bf{y}},{\bf{\hat y}})$}
	\For{$i \gets 1$ \textbf{to} $k$} {
		\If{$i\ne 1$}{
			recipdiag = $1/{\bf{R_{x}}}[i][i]$; \\
			eta[$i$] = recipdiag; \\
			\For{$p \gets i\!+\!1$ \textbf{to} $k$}{
				${\bf{R}}_{t}[i][p]={\bf{R_{x}}}[i][p]\ *$ recipdiag;
			}
		}		
		\For{$j \gets i\!+\!1$ \textbf{to} $k$}{
			temp = ${\bf{R_{x}}}[i][j]$; \\
			\For{$p \gets j$ \textbf{to} $k$}{
				${\bf{R_{x}}}[j][p] = {\bf{R_{x}}}[j][p]-{\bf{R}}_t[i][p]*$temp;
			}
		}	
	}
	$\omega^2 ({\bf{y}},{\bf{\hat y}})=1-\rho_{1{\bf{y}}}*\rho_{1{\bf{y}}}$; \\
	b$[1] = \rho_{1{\bf{y}}}$; \\
	\For{$i \gets 2$ \textbf{to} $k$}{
		tempr = $\rho_{i{\bf{y}}}$; \\
		\For{$j \gets 1$ \textbf{to} $i-1$}{
			tempr = tempr$ - {\rm{b}}[j] * {\bf{R}}_t[j][i]$;
		}
		b$[i]$ = tempr; \\
		$\omega^2 ({\bf{y}},{\bf{\hat y}})=\omega^2 ({\bf{y}},{\bf{\hat y}})-{\rm{tempr}}*{\rm{tempr}}*{\rm{eta}}[i]$;
	}
	\Return{$\omega^2 ({\bf{y}},{\bf{\hat y}})$}
	\caption{Calculation of $\omega^2 ({\bf{y}},{\bf{\hat y}})$}
	\label{alg:a2}
\end{algorithm}

\begin{table*}[!htbp]
	\centering
	\caption{Numbers of different operations for only one responsor in the proposed method and the hat-matrix method. The dimension of the vectors is $d$, and the sparsity parameter is $k$.}\label{t1}
	\begin{spacing}{1.29}
		\begin{tabular}{c p{5cm}<{\centering} p{5cm}<{\centering} c}
			\hline
			&  $+$  &  $\times$  &  $\div$  \\
			\hline
			Algorithm~\ref{alg:a1}    &  $\frac{1}{6}{k^3}+\frac{1}{2}{k^2}+\frac{1}{3}k$
			&  $\frac{1}{6}{k^3} + {k^2} + \frac{5}{6}k$  &  $k$ \\
			Algorithm~\ref{alg:a2}    &  $\frac{1}{6}{k^3}+\frac{1}{2}{k^2}+\frac{1}{3}k$  
			&  $\frac{1}{6}{k^3} + {k^2} - \frac{1}{6}k$  &  $k-1$ \\
			Hat-matrix & $(k + 3)d + \frac{1}{6}{k^3} + \frac{3}{2}{k^2} + \frac{4}{3}k$  &  $(k + 2)d + \frac{1}{6}{k^3} + 2{k^2} + \frac{17}{6}k+1$  &	 $k+1$	 \\
			\hline
		\end{tabular}
	\end{spacing}
\end{table*}

\vspace{0.2 cm}

{\center\bf{4. \quad TIME COMPLEXITY ANALYSIS}

}

Here we analyze and compare the time complexity of the proposed method and the hat-matrix method.

\vspace{0.2 cm}
{\noindent\bf{4.1 Subset Selection with Only One Responsor}
}

From Algorithms \ref{alg:a1} and \ref{alg:a2}, we can easily obtain the numbers of additions, multiplications, and divisions that need to be executed for each candidate subset in the proposed method. In the hat-matrix method, we also firstly compute all the inner products between two predictors and all the inner products between the responsor and each predictor. In the process of subset selection, the calculated inner products can be directly used to construct ${{\bf{X}}^{{T}}}{\bf{X}}$ and ${{\bf{X}}^{{T}}}{{\bf{y}}}$. Subsequently, we compute $\hat{\bm{\beta}} = {({{\bf{X}}^{{T}}}{\bf{X}})^{ - 1}}({{\bf{X}}^{{T}}}{{\bf{y}}})$ by Gauss Elimination. Lastly, ${\emph{\textbf{e}}} = {\bf{y}} - {\bf{X}}\hat{\bm{\beta}}$ and the $l_2$ norm of ${\emph{\textbf{e}}}$ are calculated for selecting the best subset. 

The numbers of different operations in the proposed Algorithms \ref{alg:a1} and \ref{alg:a2}, and the hat-matrix method are provided in Table~\ref{t1}, from which we can see that the time complexity of the proposed method is $O(\frac{1}{6}{k^3})$, and the time complexity of hat-matrix method is $O(\frac{1}{6}{k^3}+kd)$. Because the dimension $d$ of the variables in regression is generally the number of observations or experiments and large
enough, the proposed method can greatly improve the efficiency of subset selection with only one responsor.

According to Table~\ref{t1}, Algorithms \ref{alg:a1} and \ref{alg:a2} have the same time complexity if 1 division and $k$ multiplications are reduced in Algorithm \ref{alg:a1} by considering the first element 1 in correlation matrix. Hence, Algorithm \ref{alg:a2} is exactly combined with two parts decomposed from Algorithm \ref{alg:a1}: one part is completely unrelated to responsor and the other part is related to responsor. For the cases with multiple responsors, Algorithm \ref{alg:a2} is obviously more efficient than Algorithm \ref{alg:a1} because the part unrelated to responsor can be executed only once. Therefore, we can use Algorithm \ref{alg:a2} in all cases.

\vspace{0.2 cm}
{\noindent\bf{4.2 Subset Selection with Multiple Responsors}
}

According to the above analysis, for sparse regression with $m$ responsors ($m>1$), we can execute the codes from Lines 1 to 15 in Algorithm \ref{alg:a2} only once, and the codes from Lines 16 to 25 in Algorithm \ref{alg:a2} need to be executed $m$ times.

In the hat-matrix method, we let ${\bf{Y}}=[{{\bf{y}}_1},{{\bf{y}}_2}, \cdots ,{{\bf{y}}_m}]$, and then ${{\bf{X}}^T}{\bf{X}}$ and ${{\bf{X}}^T}{\bf{Y}}$ can be directly constructed by the calculated inner products. Let ${\bf{\hat y}}_i$ be the estimated vector of ${\bf{y}}_i$, $i=1,2,\cdots,m$, and let ${\bf{\hat {\bf{Y}}}}=[{{\bf{\hat y}}_1},{{\bf{\hat y}}_2}, \cdots ,{{\bf{\hat y}}_m}]$. We have two different methods to compute ${\bf{\hat {\bf{Y}}}}$ for ${\bf{Y}}$, which are shown in Eq.~(\ref{hat3}). In the first method, the order of priority is ${{({{\bf{X}}^T}{\bf{X}})}^{ - 1}}({{\bf{X}}^T}{\bf{Y}})$, and ${\bf{\hat Y}}$; in the second method, the order of priority is ${\bf{X}}{{({{\bf{X}}^T}{\bf{X}})}^{ - 1}}$ and then ${\bf{\hat Y}}$. Here we compute ${{({{\bf{X}}^T}{\bf{X}})}^{ - 1}}({{\bf{X}}^T}{\bf{Y}})$ and ${\bf{X}}{{({{\bf{X}}^T}{\bf{X}})}^{ - 1}}$ by Gauss Elimination, then we have the computational complexity for different methods as shown in Table~\ref{t2}.

\begin{table*}[t]
	\centering
	\caption{Numbers of different operations for multiple responsors in the proposed method and the hat-matrix method. The dimension of the vectors is $d$, the number of responsors is $m$, and the sparsity parameter is $k$.}\label{t2}
	\begin{spacing}{1.29}
		\begin{tabular}{c p{4.3cm}<{\centering} p{4.3cm}<{\centering} c}
			\hline
			&  $+$  &  $\times$  &  $\div$  \\
			\hline
			Algorithm~\ref{alg:a2}    &  $\frac{1}{6}{k^3}-\frac{1}{6}k+m(\frac{1}{2}k^2+\frac{1}{2}k)$ 
			&  $\frac{1}{6}{k^3}+\frac{1}{2}k^2-\frac{5}{3}k+1+m(\frac{1}{2}k^2+\frac{3}{2}k-1)$  &  $k-1$ \\
			Hat-matrix (Eq.~(\ref{hat3})$a$) & $\frac{1}{6}k^3+\frac{1}{2}k^2+\frac{1}{3}k+m(kd+3d+k^2+k)$  &  $\frac{1}{6}k^3+k^2+\frac{5}{6}k+m(kd+2d+k^2+2k+1)$  &	 $k+1$	 \\
			Hat-matrix (Eq.~(\ref{hat3})$b$) & $\frac{1}{6}k^3+\frac{1}{2}k^2+\frac{1}{3}k+m(kd+3d)+(k^2+k)d$  &  $\frac{1}{6}k^3+k^2+\frac{5}{6}k+m(kd+2d)+(k^2+2k+1)d$  &	 $k+1$	 \\
			\hline
		\end{tabular}
	\end{spacing}
\end{table*}

\begin{equation}
	\begin{array}{l}
		\underbrace {[{\bf{X}}\underbrace {[{{({{\bf{X}}^T}{\bf{X}})}^{ - 1}}({{\bf{X}}^T}{\bf{Y}})]}_1]}_2\quad \underbrace {[\underbrace {[{\bf{X}}{{({{\bf{X}}^T}{\bf{X}})}^{ - 1}}]}_1({{\bf{X}}^T}{\bf{Y}})]}_2 \\ 
		\qquad \qquad \,\,\, a \qquad \qquad \qquad \qquad \qquad \, b \\
	\end{array}\label{hat3}
\end{equation}

From Table~\ref{t2} we can see that the computational complexity of the proposed method is $O(\frac{1}{2}mk^2)$, which is far less than the time complexity $O(mkd)$ of the two hat-matrix methods in Eq.~(\ref{hat3}) when $d\gg k$. There are some slight differences of computational complexity between the two hat-matrix methods. Because ${\bf{Y}}$ needs to be constructed in the method expressed in Eq.~(\ref{hat3})$a$, we use the method in Eq.~(\ref{hat3})$b$ to reduce the space complexity of algorithm in this paper. For the method in Eq.~(\ref{hat3})$b$, we calculate ${\bf{X}}{{({{\bf{X}}^T}{\bf{X}})}^{ - 1}}$ firstly, and then each ${{\bf{X}}^T}{\bf{y}}$ can be traversed by loops.

\vspace{0.2 cm}
{\noindent\bf{4.3 Discussion}}

In this subsection we discuss the reasons why the proposed method can significantly speed up the process of subset selection in sparse regression.

In fact, according to Eq. (\ref{traditionalequation}) we have
\begin{equation}
	{\emph{\textbf{e}}}
	= {\bf{y}} - {\bf{X}}{({{\bf{X}}^{{T}}}{\bf{X}})^{ - 1}}{{\bf{X}}^{{T}}}{{\bf{y}}}={\bf{y}} - {\bf{X}}\hat{\bm{\beta}}.
	\label{discuss}
\end{equation}
From Eq. (\ref{discuss}) we can see that, for each subset to be considered, the linear coefficients in the traditional sparse regression method need to be computed firstly, then ${\emph{\textbf{e}}}$ and the $l_2$ norm of ${\emph{\textbf{e}}}$ can be calculated for selecting the best subset of predictors for ${\bf{y}}$. However, according to Eq. (\ref{eq14}), we can directly find the best subset with $k$ predictors for a responsor without calculating any linear coefficients for the predictors in candidate subsets. Hence, the linear coefficients need to be computed $m{\rm{C}}_n^k$ times in the traditional sparse regression, but $m$ times in total in the proposed method.

\vspace{0.2 cm}

{\center\bf{5. \quad CONCLUSION}

}

Sparse regression is essentially a correlation problem among predictors and responsor. In this paper, we discuss sparse regression with fixed sparsity parameter $k$ from the view of multivariate correlation, and obtain the formula of conditional uncorrelation. Then the best subset of predictors for a responsor in sparse regression can be selected by computing the ratio of two determinants of correlation matrices. Based on the formula of conditional uncorrelation, we propose an efficient method of non-approximate subset selection. For sparse regression with only one responsor, the computational cost of subset selection in the traditional hat-matrix method and the proposed non-approximate method are $O(\frac{1}{6}{k^3}+kd)$ and $O(\frac{1}{6}{k^3})$, respectively. For the case with $m$ responsors ($m>1$), the computational cost of subset selection in the traditional method and the proposed method are $O(mkd)$ and $O(\frac{1}{2}m{k^2})$, respectively. Because the dimension of the variables $d$ is generally far larger than $k$, the proposed method can greatly improve the efficiency of non-approximate subset selection.

\bibliographystyle{icml2020}
\bibliography{Reference}

\end{document}